\documentclass[letterpaper]{article}
\usepackage{proceed2e}
\usepackage[margin=1in]{geometry}
\usepackage{graphicx}
\usepackage{times}
\usepackage{adjustbox}
\usepackage{amsmath}
\usepackage{amsthm}
\usepackage{amssymb}
\usepackage{graphicx}
\usepackage{natbib}

\title{Ambient Hidden Space of Generative Adversarial Networks}

\author{
Xinhan Di$^1$, 
Pengqian Yu$^2$, 
Meng Tian$^2$, 
\\ 
$^1$ Huawei Technologies Noah's Ark Lab \\
$^2$ National University of Singapore\\
\\
dixinhan@huawei.com,
yupengqian@u.nus.edu,
tianmeng@u.nus.edu
}

\begin{document}

\maketitle



\begin{abstract}

Generative adversarial models are powerful tools to model structure in complex distributions for a variety of tasks. Current techniques for learning generative models require an access to samples which have high quality, and advanced generative models are applied to generate samples from noisy training data through ambient modules. However, the modules are only practical for the output space of the generator, and their application in the hidden space is not well studied. In this paper, we extend the ambient module to the hidden space of the generator, and provide the uniqueness condition and the corresponding strategy for the ambient hidden generator in the adversarial training process. We report the practicality of the proposed method on the benchmark dataset.

\end{abstract}

\section{Introduction}

The structure in the large dataset could be represented by generative models, and this prospective is
well studied these years. The generative models learn from the training dataset, and represent a mechanism that specify a stochastic procedure to produce samples from a probability distribution. 

These has been much progress in the neural network based generative models within adversarial framework. The adversarial framework is first pioneered by the generative adversarial network (GAN) \cite{goodfellow2014generative}. In this framework, both a generator and a discriminator are trained through the adversarial training strategy where a two-player game is played in the training process. The generator learns to map a noise vector of a low-dimensional distribution (such as standard Gaussian or normal distribution) to points in a high-dimensional space. Simultaneously, a discriminator network is trained to distinguish between real and generated samples. This is achieved through setting up a min-max game between the generator and discriminator. The adversarial framework is shown to be successful in the generation of complex distributions. 

However, this generative process requires an access to a large number of full-observed samples from the desired distribution. This large-scale dataset is hard to be obtained practically. In fact, the capture of the large-scale dataset is expensive. For example, a large number of sensing images of the brain is costly to be achieved. Besides, the sensing images are very noisy in the real world. Therefore, a training strategy for obtaining a generative model directly from noisy or incomplete samples are in great demand.      

\section{Related work}

Two distinct approaches are applied to construct generative models: one is the auto-regressive generative model \cite{kingma2013auto} and the other is the adversarial generative model \cite{goodfellow2014generative}. The adversarial framework is then shown to be more powerful than the others in modeling complex data distributions \cite{bertalmio2001navier}. It is firstly represented to generate high-quality images from middle-scale datasets \cite{goodfellow2014generative}. Several advanced adversarial networks are proposed to generate high-quality images from the large-scale dataset \cite{gulrajani2017improved}. Furthermore, as the quality of  generated images is low, spectral normalization is applied for generating high-quality image from the ImageNet dataset. Similarly, self-attention mechanism is applied to allow attention-driven, long-range dependency modeling for generating both a high variety and high quality images from ImageNet \cite{zhang2018self}. These techniques are only practical when the image samples in the training dataset have high quality. However, these techniques do not work for high quality images when the training samples are noisy.          

The methods for generation of a large-scale dataset from noisy examples is well studied. The ambientGAN strategy is applied on the generation of samples from noisy dataset. In particular, an ambient function is applied to the output end of the generator and a variety of lossy measurements are applied in order to evaluate the capability of generating samples effected by different types of noise \cite{bora2018ambientgan}. Besides, through applying both the local consistency and global consistency of images, the adversarial framework is applied to complete images of arbitrary resolutions by filling-in missing regions of any shape. Both the global and local context discriminators are trained to distinguish real images from completed images. In the adversarial framework, the image completion network is trained to fool the context discriminator networks \cite{iizuka2017globally}. A different training procedure of learning a generative model is investigated, and the generative model is learned from the transition operator of a Markov chain. This chain could de-noise random noise sample and match the target distribution from the training noise. The information from the training target example is infused \cite{bordes2017learning}. However, the de-noise module is applied to the output space of the generator network. The perspective of the application for the de-noise module in the hidden space of the generator network is not well studied. The study in this paper is meaningful as it is helpful to extend the practicality of the de-noise module for the adversarial generative model. 
                
\section{Methodology}

In the adversarial framework \cite{goodfellow2014generative}, a generator's distribution $p_{x}^{g}$ is learned over data $x$ from a prior on the input noise variables $p_{z}(z)$. The generator represents a mapping to data space as $G(z,\theta_{g})$, where $G$ denotes a function represented by a neural network with parameters $\theta_{g}$. Another neural network $D(x;\theta_{d})$ is built to output a single scalar. $D(x)$ is used to represent the probability that $x$ comes from the data rather than $p_{g}$. During the training process, $D$ is trained to maximize the probability of assigning the correct label to both the training samples and samples form $G$. Specifically, $D$ and $G$ could be seen as a two-player minimax game with the value function $V(G,D)$: $\min\limits_{G}\max\limits_{D}V(D,G)=\mathbb{E}_{x \sim p_{data}(x)}[\log D(x)] + \mathbb{E}_{z \sim p_{z}(z)}[\log(1-D(G(z)))]$.

\subsection{Ambient generative models}

Let  $r$ denote real or true distribution,  $g$ denote the generated distribution, $x$ denote the underlying space and $y$ denote the measurements.  $p_{x}^{r}$ is a real underlying distribution over $\mathbb{R}^{n}$. Lossy measurements could be observed on samples from $p_{x}^{r}$. Let $m$ be the size of each observed measurement. Parameterized by $\theta$, the lossy function is $f_{\theta}: \mathbb{R}^{n} \to \mathbb{R}^{m}$. Therefore, for a given $x$ and $\theta$, the measurements are given by $y=f_{\theta}(x)$, and the distribution over the measurements $y$ is naturally induced as $p_{y}^{r}$. That is, $Y=f_{\Theta}(X) \sim p_{y}^{r}$ if $X \sim p_{x}^{r}$ and $\Theta \sim p_{\theta}$.

In order to create an implicit generative model of $p_{x}^{r}$ given an unknown distribution $p_{x}^{r}$ and a known distribution $p_{\theta}$, the IID realizations $\{y_{1},y_{2},...,y_{s}\}$ from the distribution $p_{y}^{r}$ could be applied. However, unlike the standard GAN setting, the desired objects $X \sim p_{x}^{r}$ is not obtained. Instead, a dataset of measurements $Y \sim p_{y}^{r}$ is achieved. That is, the goal is to generate clean samples $X \sim p_{x}^{r}$ with only noisy available samples $Y \sim p_{y}^{r}$. The adversarial framework is then updated by $\min\limits_{G}\max\limits_{D}V(D,G)=\mathbb{E}_{Y^{r} \sim p_{y}^{r}}[\log D(y)] + \mathbb{E}_{z \sim p_{z}(z), \Theta \sim p_{\theta}}[\log(1-D(f_{\Theta}(G(z))))]$.

If these is a unique distribution $p_{x}^{r}$ that can induce the measurement distribution $p_{y}^{r}$, the minimax game can exist. If the discriminator $D$ is optimal such that $D(.)=\frac{p_{y}^{r}(.)}{p_{y}^{r}(.)+p_{y}^{g}(.)}$,  a generator $G$ is optimal if and only if $p_{x}^{g}=p_{x}^{r}$.

\subsection{Ambient hidden space}

\begin{figure*}[h!]
\centering
   \includegraphics[width=0.40\linewidth]{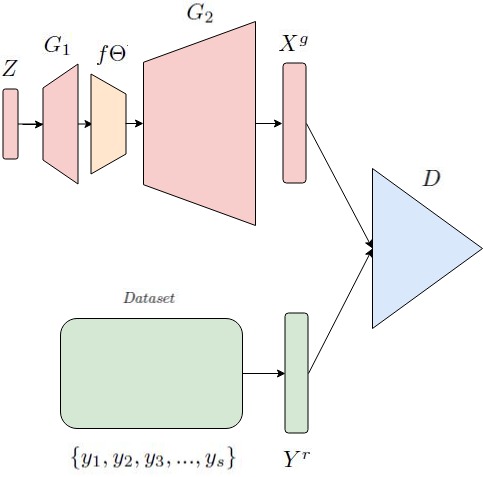}
   \caption{Hidden AmbientGAN training.}
\label{fig:1}
\end{figure*}

The ambient mapping function $f_{\Theta}$ works in the domain of $X \sim p_{x}^{r}$ or the domain of output of the generator $G$. Both domains are  in the same rgb image space. However, it is not clear whether the ambient mapping function $f_{\Theta}$ works in the hidden space of the generator $G$. We investigate the updated formulation: $\min\limits_{G}\max\limits_{D}V(D,G)=\mathbb{E}_{Y^{r} \sim p_{y}^{r}}[\log D(y)] + \mathbb{E}_{z \sim p_{z}(z), \Theta \sim p_{\theta}}[\log(1-D(G_1(f_{\Theta}(G_2(z)))))]$. That is, one $f_{\Theta}$ function mapping from $\mathbb{R}^{n}$ to $\mathbb{R}^{m}$ works in the rgb image space, and the other $f_{\Theta}$ function works in the hidden space of generator networks. In the following, we study the minimax two-player game for this setting as illustrated in Figure \ref{fig:1}.

\begin{figure*}[h!]
   \includegraphics[width=1.0\linewidth]{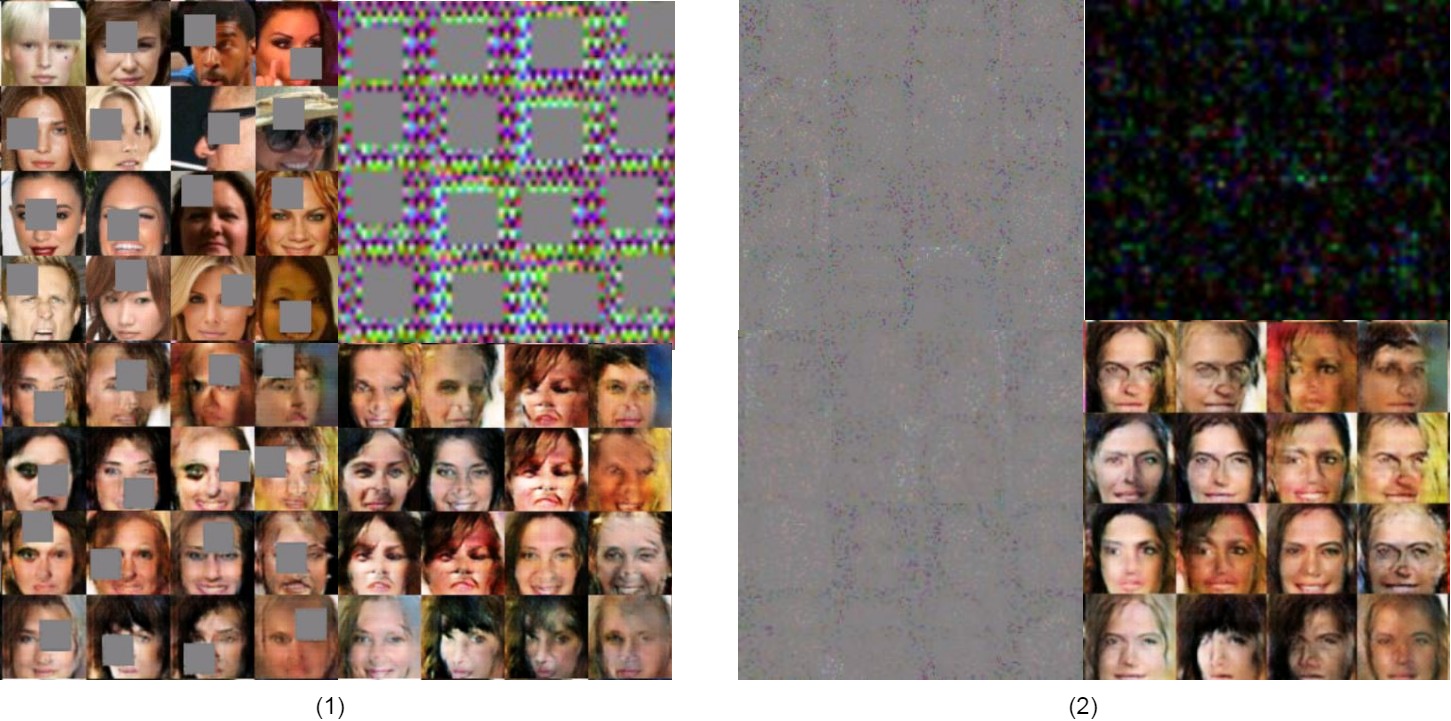}
   \caption{Block-Pixel \& Block-Patch.}
\label{fig:2}
\end{figure*}

\begin{figure*}[h!]
   \includegraphics[width=1.0\linewidth]{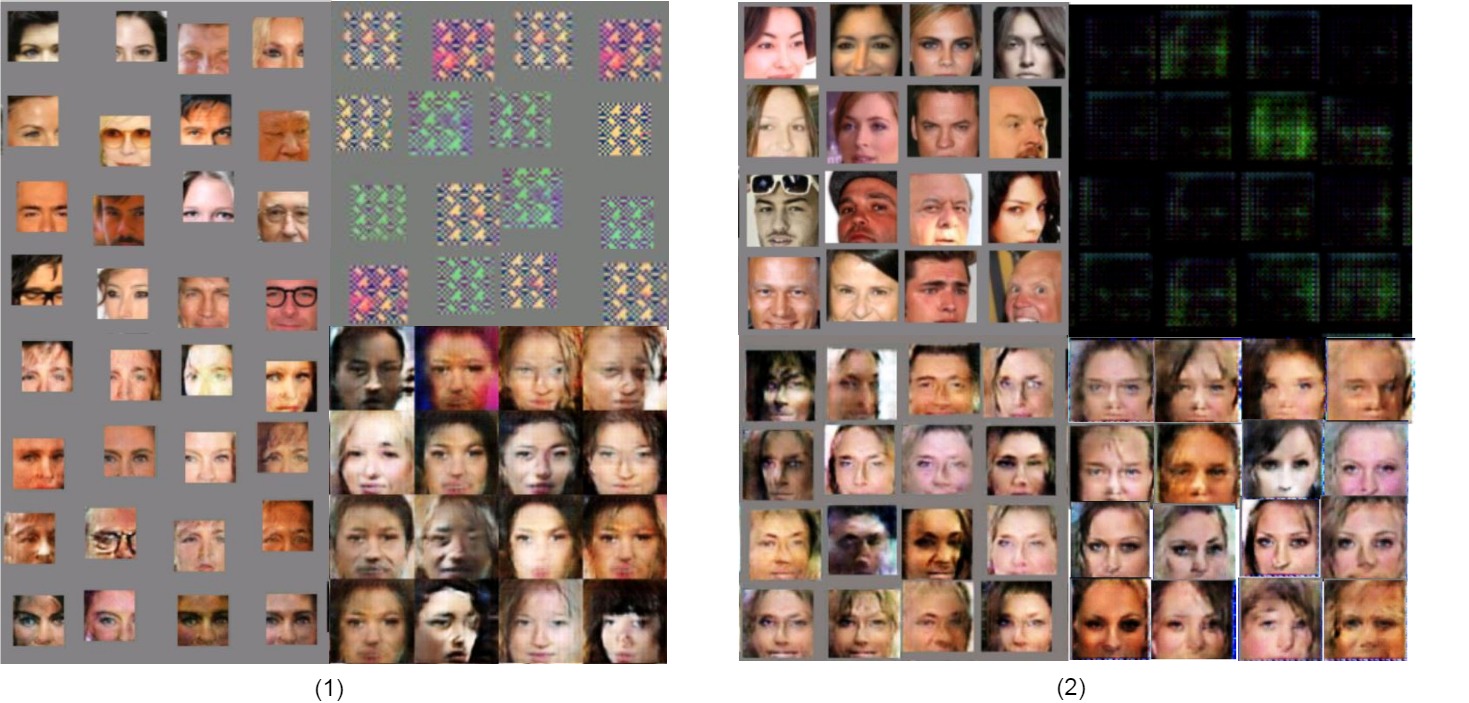}
   \caption{Keep-Patch \& Extract-Patch.}
\label{fig:3}
\end{figure*}

\subsubsection{Uniqueness}

For the $f_{\Theta}$ function at the output end of the generator $G$ and in the hidden space of $G$, we denote $l_1=f_{\Theta}(l_2)$ as the mapping process. Particularly, 
\begin{equation*}
\left\{  
    \begin{array}{lr}  
    l_1 = l_2 \quad w.p.\quad p ,&\\ 
    l_1 \neq l_2 \quad w.p.\quad 1-p.\\  
    \end{array}  
\right.  
\end{equation*}

We call this condition as the uniqueness condition for the ambient function when $p=0$. Specifically, if $l_2$ is represented as $X \sim p_{x}^{r}$, then $l_1$ is represented as $Y \sim p_{y}^{r}$. If $l_2$ is represented as a sample of the hidden space $X_h \sim p(x_h^r)$, $l_1$ is then represented as a sample of the corresponding space that is ambient $Y_h \sim p(y_{h}^{r})$ where $x_h$ and $y_h$ denote the hidden space of the generator $G$.

\subsubsection{Global optimality}

The global optimality could be achieved when the condition of the above independence is satisfied. For any given generator $G$, we first discuss the optimal discriminator $D$ as following.

The training criterion for the discriminator $D$, given any generator $G$, is to maximize the quantity $V(G,D)=\int_{y}p_{y}^{r}(y)\log(D(y))dy + \int_{z}p_{z}(z)[\log(1-D(G_{1}(f_{\Theta}(G_{2}(z)))))]dz$.

As $y=f_{\Theta}(x)$,
\begin{equation*}
\left\{  
    \begin{array}{lr}  
    y = x \quad w.p.\quad p_1 ,&\\ 
    y \neq x \quad w.p.\quad 1-p_1,\\  
    \end{array}  
\right.  
\end{equation*}
and $y_h = f_{\Theta}(x_h)$, the generator is updated as $G_1(y_h)$ when $G_2(z)$ is denoted as $x_h$. Similarly as above, $y_h=x_h$ with probability $p_2$ and  $y \neq x$ with probability $1-p_2$. That is, $G_1(f_{\Theta}(G_2(z)))=G_1(G_2(z))$ with probability $p_2$, $G_1(f_{\Theta}(G_2(z))) \neq G_1(G_2(z))$ with probability $1-p_2$. 

If $p_1=0$ and $p_2=0$, due to the Lemma 5.1 in the original ambient adversarial networks \cite{bora2018ambientgan}, both the generator $G$ and the discriminator $D$ could achieve optimal points. We denote these two points as $D_1(\cdot)=\frac{p_{y}^{r}(\cdot)}{p_{y}^{r}(\cdot)+p_{y}^{g}(\cdot)}$ and $G_1(\cdot)$ is attained if and only if $p_{x}^{g}=p_{x}^{r}$. Here $G_1(f_{\Theta}(G_2(z))) \sim p_{y}^{g}$

If $p_1 = 0$ and $p_2 \neq 0$, both the generator and the discriminator could achieve $D_1(\cdot)$ and $G_1(\cdot)$ when $G_1(f_{\Theta}(G_2(z))) \neq G_1(G_2(z))$ with probability $1-p_2$. However, when $G_1(f_{\Theta}(G_2(z)))=G_1(G_2(z))$ with probability $p_2$, the optimal points for $G$ and $D$ are different. Following the adversarial generative network framework \cite{goodfellow2014generative}, $D_2(\cdot)=\frac{p_{x}^{r}}{p_{x}^{r}+p_{x}^{g}}$ and $G_2(\cdot)$ is achieved if and only if $p_{x}^{r}=p_{x}^{g}=p_{y}^{g}$. 

If $p_2 \neq 0$, both the generator and discriminator have the probability of achieving different stable points during training. It is hard for the generator and discriminator to achieve optimal points during the two-player training. In the following, we shall use the uniqueness condition  $p_1=0$ and $p_2=0$.

\section{Numerical experiments}

The baseline is to ignore any measurement. Samples generated by the baseline and the hidden ambient models are displayed. For each experiment, samples from the dataset of measurements $y^{r}$ available for training, samples both generated by the baselines and the hidden ambient models are presented. In addition, samples generated from the hidden space of the ambient function are provided. 

\subsection{Models}

We propose several measurement models which satisfy the uniqueness condition. There are listed as following. $\textbf{Block-Pixel}:$ each pixel is independently set to zero with probability $p$. $\textbf{Block-Patch}:$ a randomly chosen $k \times k$ patch is set to zero. $\textbf{Keep-Patch}:$ all pixels outside a randomly chosen $k \times k$ patch are set to zero. $\textbf{Extract-Patch}:$ a random $k \times k$ patch is extracted. Unlike the measurement applied in the original AmbientGAN \cite{bora2018ambientgan}, the generated samples from the measurements of $\textbf{Gaussian-Projection}$, $\textbf{Convolve+Noise}$ are messy, as these two measurements do not satisfy the uniqueness condition.   

\subsection{Results}

\subsubsection{Block-Pixel}
As shown in Figure \ref{fig:2}(1), samples of lossy measurement are at the upper left. Randomly chosen patch in each image is blocked. Samples of ambient hidden space are at the upper right. Randomly chosen patch in each feature map is blocked. Samples generated by the baselines are at the lower left. Samples generated by the proposed hidden ambient model are at the lower right. 

\subsubsection{Block-Patch}
As shown in Figure \ref{fig:2}(2), samples of lossy measurement are at the upper left. Each pixel is blocked independently with probability $p=0.95$. Samples of ambient hidden space are at the upper right. Each channel is blocked independently with probability $p=0.95$. Samples generated by the baselines are at the lower left. Samples generated by the proposed hidden ambient model are at the lower right.

\subsubsection{Keep-Patch}
As shown in Figure \ref{fig:3}(1), samples of lossy measurement are at the upper left. All pixels outside a randomly chosen patch are set to zero. Samples of ambient hidden space are at the upper right. All pixels outside a randomly chosen patch in each feature map is blocked. Samples generated by the baselines are at the lower left. Samples generated by the proposed hidden ambient model are at the lower right.

\subsubsection{Extract-Patch}
As shown in Figure \ref{fig:3}(2), samples of lossy measurement are at the upper left. A random patch is extracted. Samples of ambient hidden space are at the upper right. A random patch in each feature map is extracted. Samples generated by the baselines are at the lower left. Samples generated by the proposed hidden ambient model are at the lower right.

\section{Discussions}

In this paper, we implement the ambient hidden module in the generative network during the adversarial training process, and we achieve good quality images generated from the noisy training samples under the uniqueness condition. This work is an extension of the ambient generative networks which the ambient module is only applied in the output space of the generative networks. 

In this work, we consider several ambient functions in the hidden space of the adversarial network. In order to enlarge the practicality of hidden ambient modules in the adversarial networks, it is worthwhile to study  generative methods that make use of noisy training data without knowing the noise distribution.  

\bibliographystyle{named}
\bibliography{ijcai18}

\begin{thebibliography}{}

\bibitem[\protect\citeauthoryear{Bertalmio \bgroup \em et al.\egroup
  }{2001}]{bertalmio2001navier}
Marcelo Bertalmio, Andrea~L Bertozzi, and Guillermo Sapiro.
\newblock Navier-stokes, fluid dynamics, and image and video inpainting.
\newblock In {\em Computer Vision and Pattern Recognition, 2001. CVPR 2001.
  Proceedings of the 2001 IEEE Computer Society Conference on}, volume~1, pages
  I--I. IEEE, 2001.

\bibitem[\protect\citeauthoryear{Bora \bgroup \em et al.\egroup
  }{2018}]{bora2018ambientgan}
Ashish Bora, Eric Price, and Alexandros~G Dimakis.
\newblock Ambientgan: Generative models from lossy measurements.
\newblock In {\em International Conference on Learning Representations (ICLR)},
  2018.

\bibitem[\protect\citeauthoryear{Bordes \bgroup \em et al.\egroup
  }{2017}]{bordes2017learning}
Florian Bordes, Sina Honari, and Pascal Vincent.
\newblock Learning to generate samples from noise through infusion training.
\newblock {\em arXiv preprint arXiv:1703.06975}, 2017.

\bibitem[\protect\citeauthoryear{Goodfellow \bgroup \em et al.\egroup
  }{2014}]{goodfellow2014generative}
Ian Goodfellow, Jean Pouget-Abadie, Mehdi Mirza, Bing Xu, David Warde-Farley,
  Sherjil Ozair, Aaron Courville, and Yoshua Bengio.
\newblock Generative adversarial nets.
\newblock In {\em Advances in neural information processing systems}, pages
  2672--2680, 2014.

\bibitem[\protect\citeauthoryear{Gulrajani \bgroup \em et al.\egroup
  }{2017}]{gulrajani2017improved}
Ishaan Gulrajani, Faruk Ahmed, Martin Arjovsky, Vincent Dumoulin, and Aaron~C
  Courville.
\newblock Improved training of wasserstein gans.
\newblock In {\em Advances in Neural Information Processing Systems}, pages
  5769--5779, 2017.

\bibitem[\protect\citeauthoryear{Iizuka \bgroup \em et al.\egroup
  }{2017}]{iizuka2017globally}
Satoshi Iizuka, Edgar Simo-Serra, and Hiroshi Ishikawa.
\newblock Globally and locally consistent image completion.
\newblock {\em ACM Transactions on Graphics (TOG)}, 36(4):107, 2017.

\bibitem[\protect\citeauthoryear{Kingma and Welling}{2013}]{kingma2013auto}
Diederik~P Kingma and Max Welling.
\newblock Auto-encoding variational bayes.
\newblock {\em arXiv preprint arXiv:1312.6114}, 2013.

\bibitem[\protect\citeauthoryear{Zhang \bgroup \em et al.\egroup
  }{2018}]{zhang2018self}
Han Zhang, Ian Goodfellow, Dimitris Metaxas, and Augustus Odena.
\newblock Self-attention generative adversarial networks.
\newblock {\em arXiv preprint arXiv:1805.08318}, 2018.

\end{thebibliography}

\end{document}